\documentclass[letterpaper, 10 pt, conference]{ieeeconf}  % Comment this line out if you need a4paper

\IEEEoverridecommandlockouts                              % This command is only needed if 
\usepackage{graphics} % for pdf, bitmapped graphics files
\usepackage{epsfig} % for postscript graphics files
\usepackage{mathptmx} % assumes new font selection scheme installed
\usepackage{times} % assumes new font selection scheme installed
\usepackage{amsmath} % assumes amsmath package installed
\usepackage{amssymb}  % assumes amsmath package installed
\usepackage{makecell}
\usepackage{comment}

\title{\LARGE \bf
EdgeAI Drone for Autonomous Construction Site Demonstrator*
}

\author{Emre Girgin$^{1}$, Arda Taha Candan$^{2}$, and Coşkun Anıl Zaman$^{2}$% <-this % stops a space
\thanks{*This project has received funding from the ECSEL Joint Undertaking (JU) under grant agreement No 101007321. The JU receives support from the European Union’s Horizon 2020 research and innovation programme and France, Belgium, Czech Republic, Germany, Italy, Sweden, Switzerland, Türkiye. This study was supported by Scientific and Technological Research Council of Türkiye (TUBITAK) under the Grant Number 121N347. The authors thank TUBITAK for their support.}% <-this % stops a space
\thanks{$^{1}$Faculty of Engineering, Aerospace Engineering,
        Embry-Riddle Aeronautical University, Daytona Beach, Florida 32124, USA (Author was a researcher at TUBITAK BILGEM during the project period.)
        {\tt\small girgine@my.erau.edu}}%
\thanks{$^{2}$Robotics and Autonomous Systems Division, TUBITAK BILGEM, Gebze 41400, Türkiye
        {\tt\small arda.candan@tubitak.gov.tr}
        {\tt\small anil.zaman@tubitak.gov.tr}}%
}

\begin{document}

\maketitle
\thispagestyle{empty}
\pagestyle{empty}

%%%%%%%%%%%%%%%%%%%%%%%%%%%%%%%%%%%%%%%%%%%%%%%%%%%%%%%%%%%%%%%%%%%%%%%%%%%%%%%%
\begin{abstract}

The fields of autonomous systems and robotics are receiving considerable attention in civil applications such as construction, logistics, and firefighting. Nevertheless, the widespread adoption of these technologies is hindered by the necessity for robust processing units to run AI models. Edge-AI solutions offer considerable promise, enabling low-power, cost-effective robotics that can automate civil services, improve safety, and enhance sustainability. This paper presents a novel Edge-AI-enabled drone-based surveillance system for autonomous multi-robot operations at construction sites. Our system integrates a lightweight MCU-based object detection model within a custom-built UAV platform and a 5G-enabled multi-agent coordination infrastructure. We specifically target the real-time obstacle detection and dynamic path planning problem in construction environments, providing a comprehensive dataset specifically created for MCU-based edge applications. Field experiments demonstrate practical viability and identify optimal operational parameters, highlighting our approach's scalability and computational efficiency advantages compared to existing UAV solutions. The present and future roles of autonomous vehicles on construction sites are also discussed, as well as the effectiveness of edge-AI solutions. We share our dataset publicly at github.com/egirgin/storaige-b950

\end{abstract}

%%%%%%%%%%%%%%%%%%%%%%%%%%%%%%%%%%%%%%%%%%%%%%%%%%%%%%%%%%%%%%%%%%%%%%%%%%%%%%%%
\section{INTRODUCTION}

The civil applications of autonomous systems and robotics constitute a rapidly growing area of research, driven by both public-facing innovations and high-impact developments in labor-intensive, hazardous sectors such as construction \cite{9361167, 9145591}, logistics \cite{zhang2018large, mitterberger2022tie, 8460480}, and firefighting \cite{chen2022robot}.  While user-centric technologies often attract the most attention, the broader societal benefits lie in automating critical services and reducing human risk \cite{mu2022intelligent}. A key challenge remains the reliance on high-performance computing to support AI models; however, emerging edge-AI solutions offer a promising path forward. Integrating low-power, cost-effective processing with robotics is essential to advancing scalable, safe, and sustainable automation in civil domains.

\begin{figure}[t]
  \centering
   \includegraphics[width=0.8\linewidth]{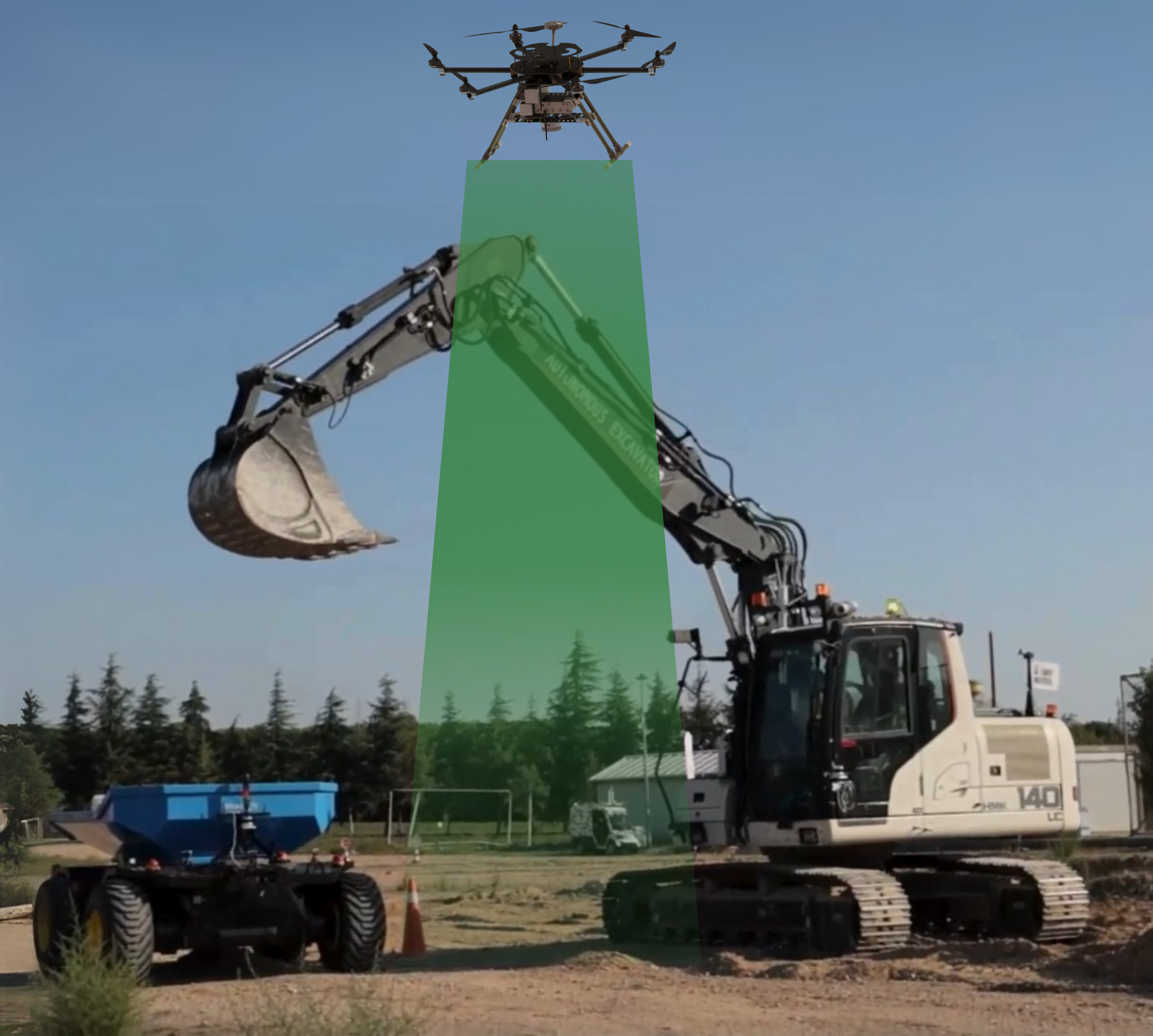}
   \caption{Our use-case demonstrates of autonomous loading in a construction site environment where the entire operation is monitored by edge-AI equipped UAV.}
   \label{fig:introduction}
\end{figure}

This paper presents a case study illustrating an autonomous operation of construction site vehicles and surveillance via AI on the edge-enabled UAV. The demonstrator includes an autonomous excavator and a custom-designed UGV while the operation site is surveyed by a custom-designed hexacopter. This hexacopter is equipped with a monocular camera and a microcontroller to detect the human presence in the operation area. All the vehicles are capable of following waypoint navigation commands autonomously that are generated by a central coordinator software that receives the obstacles and people from the drone. Additionally, all the vehicles are connected via a 5G cellular connection, which serves to provide a framework that can be generalized in the future. 

\begin{figure*}[t]
  \centering
   \includegraphics[width=1.0\linewidth]{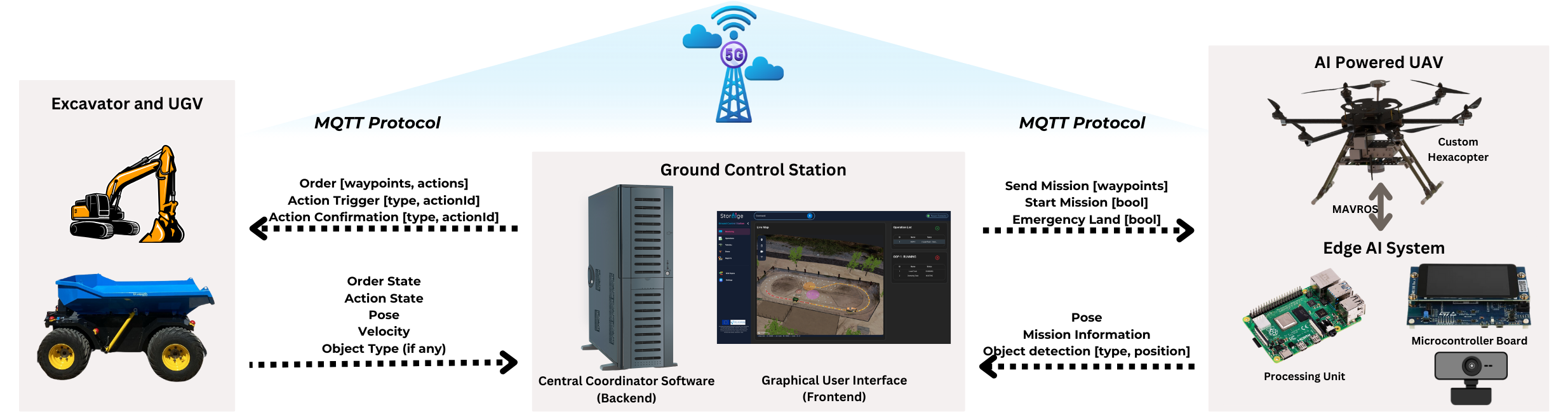}
   \caption{The overall architecture of the demonstrator is summarized. While the GCS coordinates all the vehicles, communication is facilitated by a 5G mobile base infrastructure. The UAV also contains an Edge-AI subsystem, which is responsible for detecting obstacles in the operational field.}
   \label{fig:architecture}
\end{figure*}

The custom UAV for the demonstrator features a carbon fiber-verified electro-mechanical design initially prototyped with 3D-printed components and includes a separate power system for the vision module to ensure uninterrupted processing and communication. Its software layer, built on ROS, interfaces directly with the flight control unit (FCU) to execute waypoint-based commands from the central coordinator. In order to facilitate the AI-based object detection on the microcontroller (MCU), a custom dataset has been collected and several object detection models have been trained, ensuring compatibility with the MCU. The trained model is quantized and deployed on the board. While the processing unit collects the data and conveys it to the board via serial port, the MCU processes the data with the object detection model and returns the predictions to the processing unit. This abstraction between the processing unit and the microcontroller board allows us to isolate the inference process and reduce the computational load.

Unlike previous approaches, our solution leverages lightweight AI models quantized for low-power MCUs, enabling scalable deployment at the cost minimal cost for real-time applications. In summary, we would like to highlight our contributions:
\begin{itemize}
    \item A multi-robot coordination system specifically tailored for dynamic construction site operations, integrating autonomous vehicles and UAV-based monitoring.
    \item Implementation of real-time obstacle and human presence detection using edge-AI on constrained MCU hardware, validated in a realistic construction environment governed by a central coordinator software.
    \item A novel, custom-generated dataset (TUBITAK-EdgeDrone) specifically curated for small-footprint edge AI applications in UAV-based real-time construction site surveillance.
\end{itemize}

\section{Related Work}
Despite its global economic importance, the construction industry faces persistent challenges such as low productivity growth, high labor costs, and inefficient production \cite{loveridge2017robots, debrah2022artificial}. In response, various construction robots have been explored, particularly mobile construction robots (MCRs) that combine robotic arms with mobile platforms \cite{wu2022survey, xiao2022recent, sari2022cloud}. 
%Applications span demolition \cite{mu2022intelligent}, maintenance \cite{chen2022robot} and assembly using prefabricated components like 3D-printed parts, timber, and steel \cite{zhang2018large, mitterberger2022tie, 8460480}, as well as on-site tasks such as bricklaying and painting \cite{9361167, 9145591}.
The literature also explores converting existing construction vehicles into autonomous systems \cite{borngrund2022deep}. For instance, Jin et al. \cite{jin_autonomous_2023} developed an Autonomous Excavator System (AES) for truck loading tasks \cite{ jin2023learning}, though it is limited to single-excavator control. In contrast, our system integrates multiple vehicle types, including UAVs. While UAVs are frequently used for inspection and planning purposes \cite{liang2023towards, freimuth2018planning}, our approach focuses on their real-time operational role in dynamic scenarios. Moreover, unlike common edge-AI and 5G-based UAV systems \cite{ choi2024integrating}, our demonstrator utilizes lightweight MCU-based AI for coordination with ground vehicles.

\section{Methodology}
This section presents the demonstrator’s architecture, UAV system design, and central coordination for multi-vehicle collaboration. It also covers the deployment of a custom-trained AI model on an STM32 microcontroller for low-power object detection.

\subsection{Architecture}

The demonstrator aims to execute a coordinated autonomous load-dump operation, with all entities—ground vehicles, UAV, and central coordinator—interconnected via 5G and monitored by an AI-powered STM32-equipped drone. The architecture (Figure \ref{fig:architecture}) relies on a central coordinator for global planning, while vehicles perform waypoint-based navigation and share positional data in a closed-loop setup. The UAV contributes by detecting dynamic obstacles and transmitting their global locations for collision-free path planning. Communication between vehicles and the ground control system is handled via MQTT, chosen for its lightweight, QoS-enabled messaging and seamless integration with ROS through the mqtt\_bridge, ensuring reliable interoperability across the system.

\subsection{Central Coordinator Software}

To manage coordination, we developed a Ground Control Station (GCS) that acts as a central interface between the human supervisor and the autonomous systems. The GCS oversees task distribution, monitors vehicle status, and supports real-time decision-making. It is structured with a modular architecture comprising a front-end for visualization and interaction, and a back-end for communication and task execution. This separation enhances scalability and allows independent system upgrades. The front-end, built with React and Redux, offers a responsive web interface accessible from any device and integrates Mapbox for site visualization. It communicates with ROS-based back-end components via roslibjs and mqtt\_bridge to ensure real-time interaction. The GCS back-end is built on ROS, leveraging its modular messaging framework to coordinate various subsystems. Core ROS nodes include the operation manager for task parsing, vehicle agents for direct control, a task planner to assign tasks based on proximity, and a path planner that uses Voronoi graph-based A* algorithms for safe and efficient navigation. Communication follows the VDA5050 protocol to ensure interoperability with AGV systems. By using lightweight MQTT messaging and standard-compliant formats, the GCS maintains robust communication across the 5G-connected ecosystem. This system architecture enables responsive, scalable control of multiple autonomous agents on dynamic construction sites.

\subsection{Autonomous Drone Design}

\begin{figure}[t]
  \centering
   \includegraphics[width=0.8\linewidth]{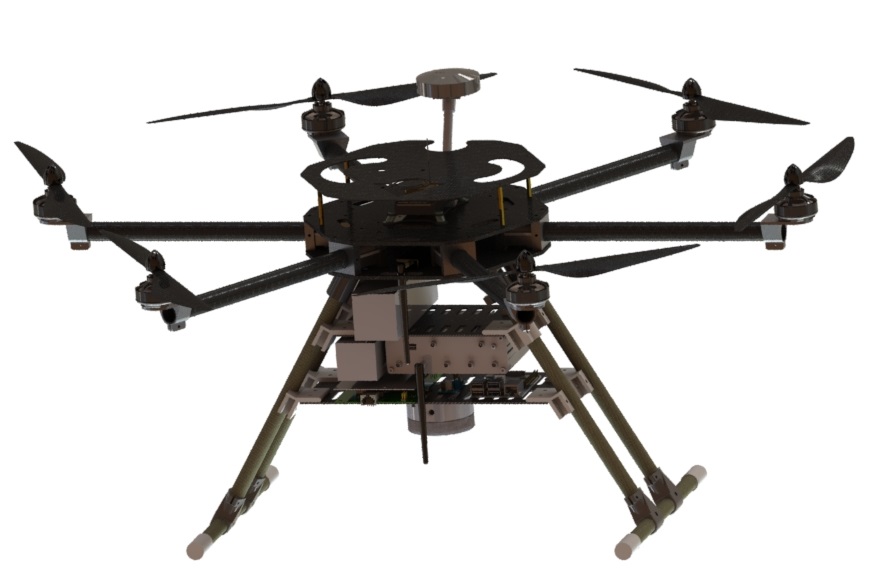}
   \caption{The CAD drawing of the hexacopter, including the AI subsystem. The carbon-fiber layers situated between the landing gear houses the batteries, RPI, MCU board, 5G router, and camera.}
   \label{fig:drone_cad}
\end{figure}

We developed a custom hexacopter equipped with a Pixhawk-based flight control unit, six BLDC motors, and a modular carbon-fiber frame designed for research and autonomous aerial surveying. (Figure \ref{fig:drone_cad}) Optimized for stable 10–11 minute flight durations, the drone integrates a general-purpose onboard computer, STM32F769i-Discovery board, a 5G router, and separate power for sensors. Real-time image capture is handled by a base-mounted camera linked to the onboard computer running ROS Noetic with MAVROS for flight control and MQTT for inter-system communication. A vision node sends camera frames to the MCU, where a quantized AI model performs object detection for subsequent real-world localization.

\subsection{AI on the Edge and Data Collection}

Deploying AI models on MCUs, presents significant limitations in terms of processing power, memory, and model complexity. Quantization techniques have emerged as a viable solution to compress models while retaining acceptable performance. In our work, we explored these methods with multiple object detection models using both a publicly available dataset \cite{zhu2018vision} and our own custom dataset, and subsequently deploying them on MCU.

To estimate the world coordinates of detected objects using a monocular top-down UAV camera—a task ill-posed for traditional structure-from-motion due to limited pose diversity—we developed a lightweight, real-time method based on the UAV’s altitude and object pixel dimensions. A RANSAC model was trained on data from a cone-marked site to learn the relationship between image scale and real-world distance. The resulting local coordinates were converted to global ones using geodesic transformations based on GNSS data, allowing the central coordinator to treat detected objects as dynamic obstacles. This pipeline, executed via quantized inference on an MCU, maintained real-time performance without overloading onboard resources.

A major bottleneck in this process was the limited input resolution supported by MCU-based models, which made existing datasets, such as VisDrone\cite{zhu2018vision}, less effective due to their low object-to-frame size ratios. To address this gap, we developed the TUBITAK-EdgeDrone dataset, specifically curated for edge AI applications. It contains approximately 25,000 training and 5,000 testing images, well-suited for small input sizes. Additionally, the dataset includes GNSS metadata to support tasks such as visual odometry, and also provides a high-resolution construction site image set for more powerful platforms. This dual-level dataset supports a wide range of applications, from lightweight MCU deployment to broader robotics and computer vision research.

\section{Experiments}

The experiments are both conducted in a laboratory setting and field tests. Field tests entail the integration of all demonstrator members and the subsequent generation of qualitative insights.

\subsection{Object Detection and Pose Estimation}

We have deployed multiple object detectors on both a regular general-purpose computer with CPU i7-8700, Raspberry PI 4 with processor Cortex-A72, and STM32H769-I with M7-Cortex. Table \ref{tab:model_fps_quant} summarizes the speed of several neural networks on different hardware. Our experiments show that YOLOLC model with input size 192x192 shows the best performance in terms of speed. While neural networks with higher input resolution had a slightly better accuracy, the difference inbetween was not significant so we did not included that into table. Note that all the deep learning networks are quantized for the MCU.

\begin{table}[h]
\caption{FPS of quantized models on different processors.}
\label{tab:model_fps_quant}
\centering
\begin{tabular}{|c||c|c|c|c|}
\hline
\makecell{\textbf{Model} \\ \textbf{(Resolution)}}&
\makecell{\textbf{Size} \\ \textbf{(KB)}} & 
\makecell{\textbf{i7-8700} \\ \textbf{(3.2 GHz)}} & 
\makecell{\textbf{A72} \\ \textbf{(1.8 GHz)}} & 
\makecell{\textbf{M7} \\ \textbf{(216 MHz)}} \\
\hline
ST-SSDv1-192 & 594.4 & 365.1 & 124.7 & 0.39 \\
\hline
ST-SSDv1-256 & 789.8 & 208.8 & 74.4 & 0.21 \\
\hline
SSDv2-256 & 1559.2 & 66.4 & 33.6 & 0.13 \\
\hline
SSDv2-416 & 1604.2 & 25.0 & 12.7 & 0.06 \\
\hline
YoloLC-192 & 339.3 & 796.6 & 257.9 & 0.85 \\
\hline
YoloLC-256 & 339.3 & 447.1 & 131.2 & 0.44 \\
\hline
TinyYolo-224 & 11105.5 & 74.7 & 18.2 & 0.13 \\
\hline
TinyYolo-416 & 11105.5 & 22.0 & 5.4 & 0.05 \\
\hline
\end{tabular}
\end{table}

The RANSAC model that we have adopted is trained to learn the relation between the number of pixels that a meter represents and the altitude of the drone. We have trained multiple models with different polynomial degrees. We then tested their performance on the unseen data, where the actual distance between two pixels is known. Table \ref{tab:ransac_results} shows the performance of the models with different complexities.

\begin{table}[h]
\caption{RMSE of RANSAC estimators in cm (\textbf{lower} is better).}
\label{tab:ransac_results}
\begin{center}
\begin{tabular}{|c||c|c|c|c|}
\hline
\textbf{Polynomial Degree} & \textbf{5 m} & \textbf{8 m} & \textbf{10 m} & \textbf{15 m} \\
\hline
Linear (1)   & 46.9 & 3.2 & 26.5 & 32.3 \\
\hline
Quadratic (2) & 12.1 & 6.9 & 25.7 & 24.1 \\
\hline
Cubic (3)     & \textbf{5.2} & 3.7 & \textbf{23.4} & \textbf{22.2} \\
\hline
Quartic (4)   & 5.6 & \textbf{2.7} & 24.2 & 29.4 \\
\hline
\end{tabular}
\end{center}
\end{table}

As model complexity increased, so did the accuracy of real-world distance estimation, with the cubic model proving sufficient for the task and thus selected for field deployment. Error rates were higher at both low altitudes—due to repeated cone detections—and high altitudes, where cones appeared smaller and harder to detect, leading to the choice of approximately 8 meters as the optimal flight altitude for testing.

\subsection{Field Tests}

\begin{figure}[h]
  \centering
   \includegraphics[width=\linewidth]{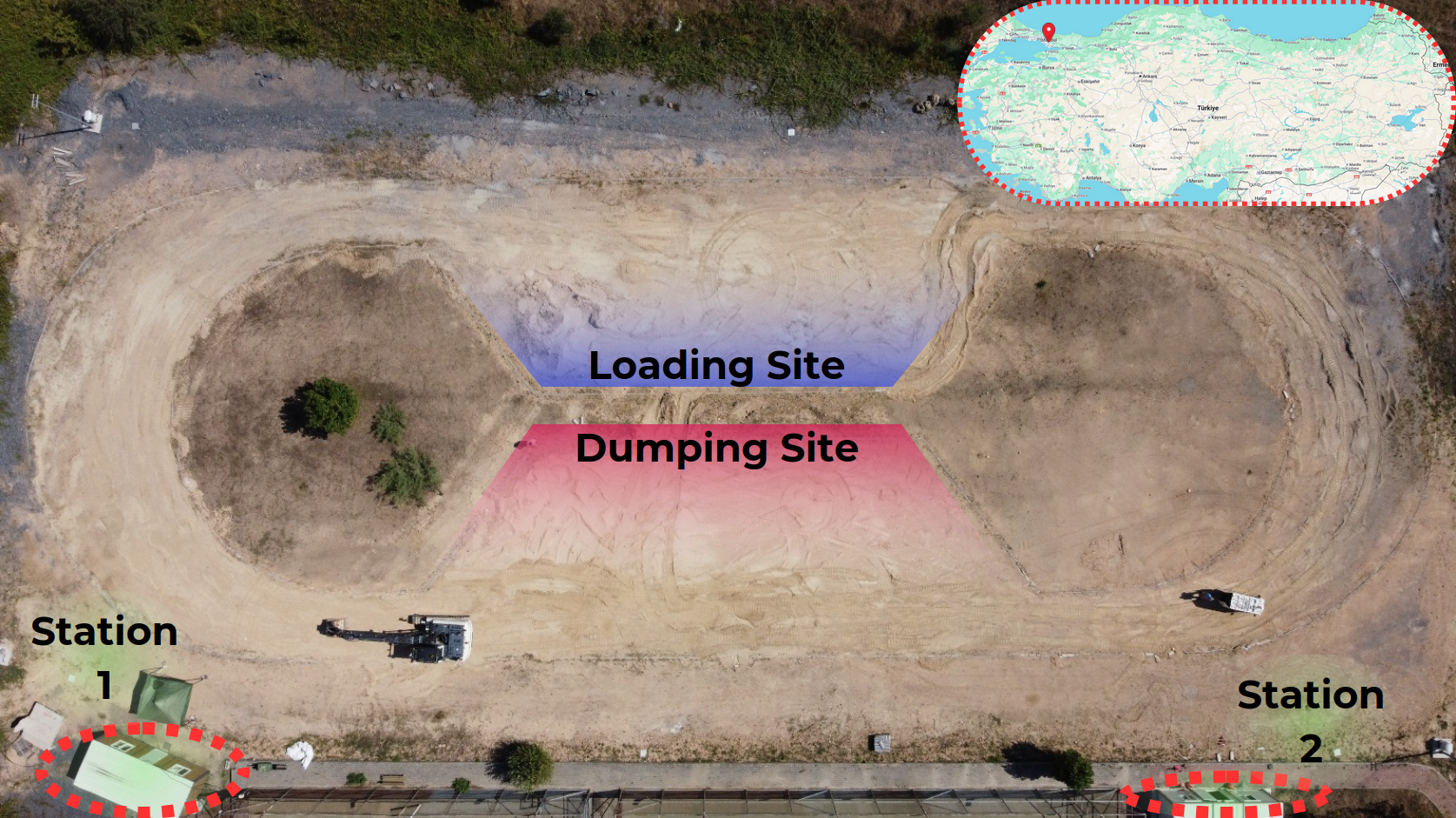}
   \caption{OpenAirLab is an autonomous construction test site in Gebze, Turkey, comprising an area of 50 meters by 100 meters.}
   \label{fig:openairlab}
\end{figure}

The field tests are conducted in the OpenAirLab, situated on the TÜBİTAK Gebze Campus, Kocaeli, Türkiye. The area, which encompasses approximately 5000 square meters, is equipped with WiFi connectivity, full coverage solid-state LIDARs for instantaneous point cloud generation, IP cameras, and 5G cellular coverage. Figure \ref{fig:openairlab} shows the OpenAirLab. 

The scenario comprises loading and dumping tasks for the autonomous excavator and a UGV. The autonomous UAV conducted an aerial survey of the area and relayed the locations of detected individuals to the GCS, which then adjusted the routes of the heavy vehicles accordingly.

The excavator and the UGV navigated for around 250 meters in total. The route they followed, loading, and dumping sites are shown in the Figure \ref{fig:openairlab}. Since the performance of the detection models are not enough to satisfy real-time requirements, as shown in the Table \ref{tab:model_fps_quant}, we used the A72 to run the AI during field tests. 

\section{Conclusion Acknowledgement}

This demonstrator presents the AI capabilities of MCUs on UAVs for construction applications. Comprising a central coordinator, a 5G connection, and an edge AI, the system shows that despite current MCU limitations, their scalable use in construction automation is promising.

We would like to express our gratitude to Büyütech for the autonomous excavator, Ford Otosan for a custom-designed UGV, and Turkcell for the 5G communication infrastructure.

\addtolength{\textheight}{-12cm}   % This command serves to balance the column lengths

%%%%%%%%%%%%%%%%%%%%%%%%%%%%%%%%%%%%%%%%%%%%%%%%%%%%%%%%%%%%%%%%%%%%%%%%%%%%%%%%

\end{document}